\documentclass[10pt,twocolumn,letterpaper]{article}

\usepackage{cvpr}              

\usepackage[dvipsnames]{xcolor}

\definecolor{cvprblue}{rgb}{0.21,0.49,0.74}
\usepackage[pagebackref,breaklinks,colorlinks,citecolor=cvprblue]{hyperref}
\usepackage{multirow}
\usepackage{amsthm}
\newtheorem{definition}{Definition}

\def\paperID{*****} 
\def\confName{CVPR}
\def\confYear{2024}

\title{CSCO: Connectivity Search of Convolutional Operators}

\author{
Tunhou Zhang\textsuperscript{\rm 1}, 
Shiyu Li\textsuperscript{\rm 1}, 
Hsin-Pai Cheng\textsuperscript{\rm 3}, 
Feng Yan\textsuperscript{\rm 2}, 
Hai Li\textsuperscript{\rm 1}, 
Yiran Chen\textsuperscript{\rm 1} 
\\
\textsuperscript{\rm 1}ECE Department, Duke University, Durham, NC 27708\\
\textsuperscript{\rm 2}Department of Computer Science, University of Houston, Houston, TX 77204\\
\textsuperscript{\rm 3} Qualcomm AI Research, San Diego, CA 92121 \\
\textsuperscript{\rm 1} \{tunhou.zhang,shiyu.li,hai.li,yiran.chen\}@duke.edu,\\
\textsuperscript{\rm 2} fyan5@central.uh.edu, \\
\textsuperscript{\rm 3} hsinpaic@qti.qualcomm.com
}

\begin{document}
\maketitle

\begin{abstract}
Exploring dense connectivity of convolutional operators establishes critical ``synapses'' to communicate feature vectors from different levels and enriches the set of transformations on Computer Vision applications.
Yet, even with heavy-machinery approaches such as Neural Architecture Search (NAS), discovering effective connectivity patterns requires tremendous efforts due to either constrained connectivity design space or a sub-optimal exploration process induced by an unconstrained search space.
In this paper, we propose CSCO, a novel paradigm that fabricates effective connectivity of convolutional operators with minimal utilization of existing design motifs and further utilizes the discovered wiring to construct high-performing ConvNets.
CSCO guides the exploration via a neural predictor as a surrogate of the ground-truth performance. 
We introduce Graph Isomorphism as data augmentation to improve sample efficiency and propose a Metropolis-Hastings Evolutionary Search (MH-ES) to evade locally optimal architectures and advance search quality.
Results on ImageNet show $\sim 0.6\%$ performance improvement over hand-crafted and NAS-crafted dense connectivity. Our code is publicly available \href{https://github.com/lordzth666/-CVPRW-CSCO}{here}.
\end{abstract}

\section{Introduction}
The fundamental success of Convolutional Neural Network (CNN) on Computer Vision lies in the effective wiring pattern, represented by dense connectivity within convolutional layers~\cite{huang2017densely,huang2018condensenet} and atomic-level neurons~\cite{chedotal2010wiring}.
Throughout neural synapses, a convolution operator, as an elementary atomic building operator, establishes receptive fields to extract spatial-local information in 2D images.

However, from classic CNNs~\cite{NIPS2012_c399862d,he2016deep,simonyan2014very} to modernized CNNs driven by Neural Architecture Search (NAS)~\cite{tan2019efficientnet,tan2019mnasnet}, the construction of CNNs mainly innovates an effective building block composed of a combination of building operators and directly stacks a few copies of these operators to construct the overall architecture.
On images, most CNN designs are constrained to a chain-like structure without delicate consideration of building block connectivity. 
On the one hand, hardware is designed to handle better chain-like architectures such as MobileNets~\cite{howard2017mobilenets,sandler2018mobilenetv2,howard2019searching}.
On the other hand, chain-like CNN architectures are easier to study, requiring less extensive efforts to fully explore, thus yielding a better rate of improvement (ROI) in research and development on vision benchmarks.
The limitations in the aforementioned chain-like designs may prevent the discovery of effective inter-block synapses that enhance feature interaction in different positions of CNN architectures.

As a result, more recent works start to scratch the surface of dense connectivity by constructing a graph representation of the network design space~\cite{zoph2018learning,pham2018efficient,liu2018darts,wortsman2019discovering}. These works explicitly seek a building cell with searchable wiring of building operators via Directed Acyclic Graphs (DAGs) as the design motif for CNN architectures.
Various search strategies are implemented to achieve a good architecture outcome, such as differentiable-based search~\cite{liu2018darts,liang2019darts+}, Bayesian Optimization~\cite{kandasamy2018neural,white2021bananas}, and local search~\cite{den2021local,white2020local}.
However, these methods employ brutal-force optimization algorithms such as differentiable-based search~\cite{liu2018darts,huang2020explicitly}, reinforcement-learning~\cite{zoph2018learning,pham2018efficient} without any topology consideration when seeking the optimal wiring of building operators.
Despite the remarkable success, existing methods may have the following challenges: (1) The constrained design space does not support the dense connectivity of versatile building operators that integrate feature vectors from all building cell levels without limitations. (2) With unconstrained dense connectivity, exploring such design space is difficult without topological information in the graph representation of building cells, such as isomorphism in adjacency matrices and locality contained with similar graphs.

In this paper, we tackle the above challenges by proposing a new paradigm, CSCO (\textbf{C}onnectivity \textbf{Search} of \textbf{C}onvolutional \textbf{O}perators), that enables the delicate exploration of the structural wiring within building cells for CNN architectures.
CSCO establishes a hierarchical structure of CNN architectures via a meta-graph comprising several Directed Acyclic Graphs (DAGs).
Each DAG represents a building cell in each hierarchy.
Within each building cell, CSCO integrates a structural design space with versatile building operators (e.g., convolution, depthwise convolution) with varying transformation capacities of input features, allowing dense connectivity searches.
The combination of dense edge connectivity and versatile heterogeneous operators we employ can craft various design motifs. 
For example, depthwise separable convolution~\cite{howard2017mobilenets}, Inverted Bottleneck~\cite{sandler2018mobilenetv2}, Inception-like~\cite{Szegedy_inceptionv4}, and Condensenets~\cite{huang2018condensenet}, etc.
This covers most design motifs from hand-crafted CNN design principles and more recent block-based search spaces in NAS (e.g., ProxylessNAS~\cite{cai2018proxylessnas}, MobileNetV3~\cite{howard2019searching} etc.), providing more opportunities to obtain top-performing CNN architectures with minimal design space constraints and priors.

Intuited by predictor-based NAS~\cite{wen2020neural,dudziak2020brp} that accurately models the design space via a surrogate model of the ground-truth performance, we follow this principle and address two key factors to demystify search on dense connectivity. 
First, we propose Graph Isomorphism to enrich architecture-performance pairs during the sampling phase of predictor-based NAS, enhancing the quality of performance prediction with improved sample efficiency. 
As a result, Graph Isomorphism advances prediction reliability in dense connectivity design space with large cardinality.
Second, we propose Metropolis-Hastings Evolutionary Search (MH-ES), which evades local optimal solutions during search space exploration.
This allows us to approach a better region of the dense connectivity design space and discover better building cells for CNNs.

CSCO improves both the evaluation strategy (i.e., the quality and reliability of prediction) and the search strategy (i.e., the quality of top-performing architectures discovered) in predictor-based NAS. As a result, CSCO discovers good CNN architectures that achieve impressive empirical results over existing hand-crafted and NAS-crafted connectivity on ImageNet.
We summarize our contributions as follows:
\begin{itemize}[noitemsep,leftmargin=*]
    \item We propose a new paradigm, CSCO, to automatically explore dense connectivity within building cells to fabricate CNN architectures. CSCO supports dense connectivity search on structural wiring of versatile convolutional building operators to seek the optimal CNN architecture.
    \item We pioneer using predictor-based NAS in dense connectivity search and demonstrate two essential techniques that advance search quality and efficiency. Specifically, we propose Graph Isomorphism to improve sample efficiency and introduce Metropolis-Hastings Evolutionary Search (MH-ES) to improve search quality.
    \item We thoroughly evaluate each component of CSCO and demonstrate 0.6\% accuracy gain over existing NAS-crafted dense connectivity CNN architectures under mobile computation regimes.
\end{itemize}

\section{Related Work} 

\noindent \textbf{Dense Connectivity of Convolutional Neural Networks.}
Existing NAS methods emphasize cell design on a graph design space~\cite{liu2018darts,pham2018efficient}; these methods are primarily topology-agnostic.
For example, DARTS implies addressing the bi-level optimization problem without considering any graph information (e.g., graph adjacency and graph isomorphism).
More recent works~\cite{wortsman2019discovering,fang2020densely,nasgem_aaai21,zhang2019autoshrink} manage to incorporate topological information into search space design and improve the flexibility of crafted CNN architectures. Yet, they still focus on a constrained search space emphasizing either a single-cell design (i.e., normal cell and./or reduction cell) or macro-level connections between building cells with constraints and limitations.
Another line of research takes advantage of network generators~\cite{Xie_2019_ICCV,ru2020neural,xie2021understanding} to obtain the best cluster of CNN architectures. Yet, these methods emphasize discovering the top-performing local regions of the search space and may miss the opportunity to discover an individual architecture with a globally optimal solution.

\noindent \textbf{Predictor-based NAS.}
Neural predictors~\cite{wen2020neural} are potent tools to map candidate architectures to their performance in a search space. 
Predictor-based NAS has two significant phases: (1) train an accuracy predictor based on exploitable architecture-performance samples collected from a search space, and (2) utilize the accuracy predictor to probe the whole search space and obtain the top-performing architectures.
Existing works enhance predictor-based NAS in sample-efficiency~\cite{dudziak2020brp,liu2021homogeneous}, quality of prediction~\cite{dai2021fbnetv3}, and better regions of the search space~\cite{wu2021stronger}. 
Yet, these works focus on a constrained block-based search space without considering topology, making them unsuitable for exploring cell structures for better sample efficiency.
In addition, existing approaches mostly employ Evolutionary Search~\cite{pincus1970letter,deb2002fast,real2019regularized} as the backbone methodology to obtain top-performing architectures on neural predictors, gradually approaching a narrower local region of the dense connectivity design space and ending up with locally optimal architectures.

\begin{figure*}[t]
\begin{center}
    \includegraphics[width=1.0\linewidth]{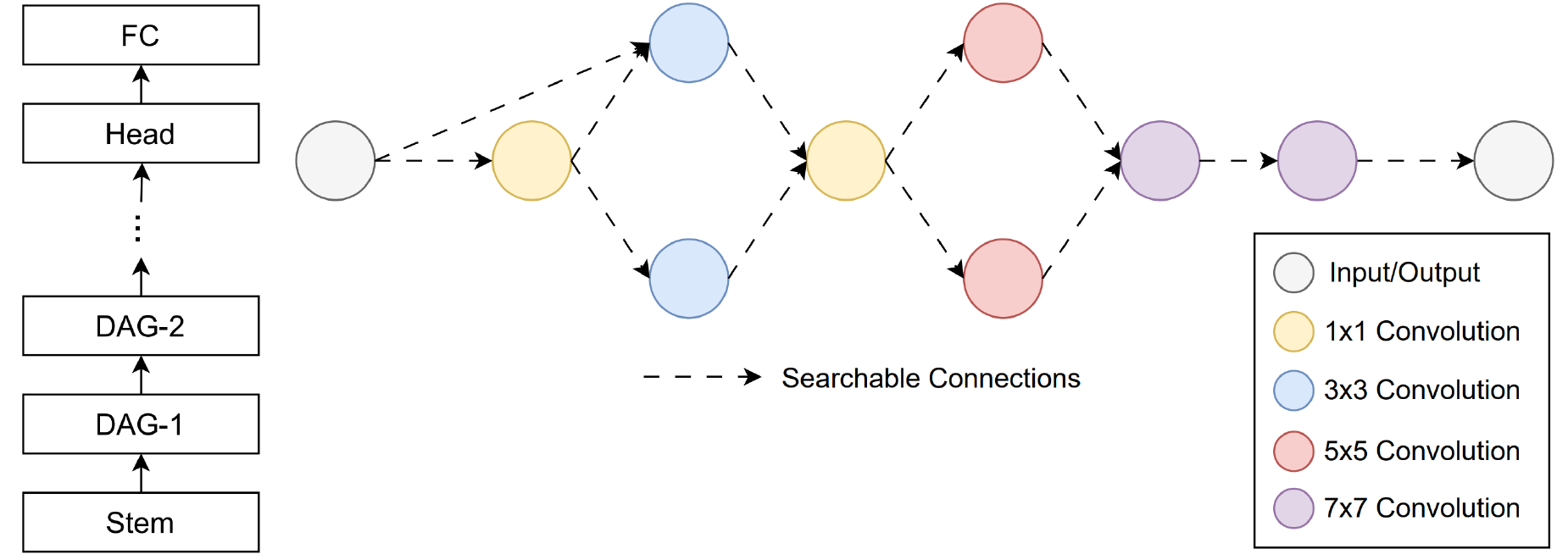}
    \caption{Overview of dense connectivity design space.}
    \label{fig:wiring_space}
\end{center}
\vspace{-1em}
\end{figure*}

\section{Dense Connectivity Design Space}
CNN architectures comprise a series of building cells that transform input features into learned representations.
For input features with varying properties (e.g., different spatial sizes), CNN architectures employ stage to represent the building cells with identical input sizes and repeat building cells to build deeper architectures.
Our dense connectivity design space delicately seeks the wiring of versatile convolutional operators in a building cell.
Figure \ref{fig:wiring_space} demonstrates an overview of our dense connectivity design space.
We utilize existing positional settings (i.e., $\times D$, $C$ denotes $D$ copies of DAG with $C$ base channels) and delicately seek the edge connections that wire versatile building operators independently for all building cells. 
We first discuss the graph representation of CNN architectures in dense connectivity design space and then discuss versatile building operators. 

\subsection{Graph Representation of CNN}
Given a CNN architecture with $K$ stages, we specify independent Directed Acyclic Graphs (DAG) for each hierarchical level of input features to represent the dense connectivity of building versatile operators.
A DAG is a building cell with multiple CNN layers and identical spatial feature sizes (i.e., image height and width).
Each DAG $\mathcal{G}^{(k)}=(\mathcal{V}^{(k)}, \mathcal{E}^{(k)})$ contains $N$ vertices and $[N\cdot(N-1)/2]$ edges with total capacity for dense connectivity search.
Among $N$ vertices, vertex 0 is defined as the input vertex that receives an output from the previous building cell, and vertex $N$ is the output vertex that sends an output to a succeeding building cell.
We define vertex $1\sim N-1$ as intermediate vertices. 
Each intermediate vertex takes input features $X$ from an arbitrary number of preceding vertices and produces an output $Y$ via an assigned building operator $op$.
More specifically, each intermediate vertex $v$ can make a connection to any preceding vertex $u_0, ..., u_{M}$, concatenates all these inputs, and use the assigned building operator $op$ to produce output representations as follows:
\begin{equation}
    Y_v = op_{v}[Concat(X_{u_0}, X_{u_2}, ..., X_{u_M})],
\end{equation}
$Concat$ denotes the feature concatenation in the channel dimension, and $M$ denotes the number of connections that vertex $v$ makes to preceding vertices.

The output vertex collects the output representations from each leaf vertex in DAG and concatenates them as an output to the succeeding building cells as follows:
\begin{equation}
    Y_{N}=Concat(\{Y_{u_i} | d_{out}(i) = 0\}),
\end{equation}
where $d_{out}(\cdot)$ denotes the out degree of a vertex. A meta-graph $\mathbf{G} = (\mathcal{G}^{(1)}, ..., \mathcal{G}^{(K)}) = (\mathbf{V}, \mathbf{E})$ composes $K$ DAGs to build the overall CNN architecture that processes input features of different hierarchies.
We construct each candidate architecture $\mathcal{A}$ by a function of vertices and edge connections on the meta-graph (i.e., the union of all independent DAGs): $\mathcal{A} = f_{arch}(\hat{\mathcal{V}}^{(1)}, ..., \hat{\mathcal{V}}^{(K)}, \hat{\mathcal{E}}^{(1)}, ...,  \hat{\mathcal{E}}^{(K)}; \hat{op}^{(1)}, ..., \hat{op}^{(K)})$. In dense connectivity design space, we assign each vertex an atomic convolutional operator from a set of versatile building operators and seek the best connectivity by optimizing edge connectivity $\hat{\mathcal{E}}^{*}$ as follows:
\begin{equation}
\tiny
\arg \max _{\mathcal{E}^{*} \subset \textbf{E}} Perf(f_{arch}(\hat{\mathcal{V}}^{(1)}, ..., \hat{\mathcal{V}}^{(K)}, \mathcal{E}^{*(1)}, ...,  \mathcal{E}^{*(K)}; \hat{op}^{(1)}, ..., \hat{op}^{(K)})),
\end{equation}

where $Perf(\cdot)$ denotes the performance metric, and $f_{arch}$ transforms a meta-graph representation to a concrete CNN architecture. Given a meta-graph with $K$ independent DAGs (e.g., $K$ stages) and $N$ vertices each, a dense connectivity design space optimizes $O(K \cdot N^2)$ dense edge connections to seek the optimal architecture and contains a much richer source of architecture fabrications.

\begin{figure*}[t]
    \begin{center}
    \includegraphics[width=\linewidth]{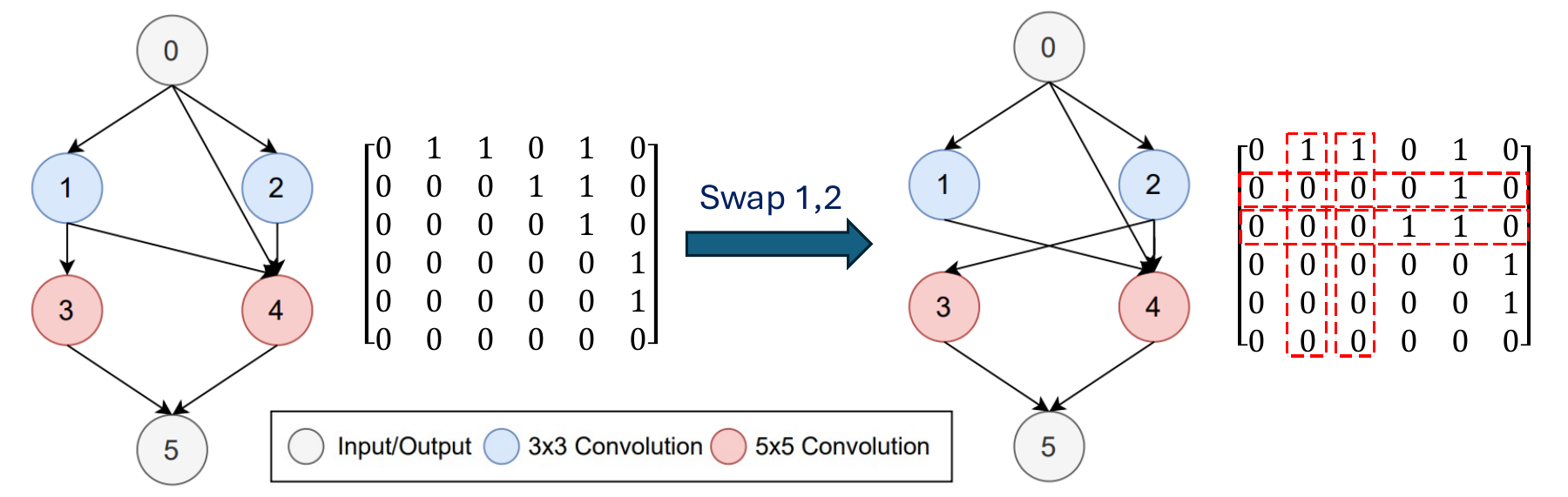}
    \caption{Graph Isomorphism creates extra training examples without extra cost.}
    \label{fig:graph_aug}
    \end{center}
\end{figure*}

\subsection{Convolution Building Operators}
\label{sec:building_op}
The dense connectivity design space is built upon versatile atomic building operators to remove the constraints in CNN design and enhance flexibility during the search.
An atomic convolution operator should contain precisely one convolution operation, followed by batch normalization~\cite{ioffe2015batch} and ReLU activation~\cite{simonyan2014very}.
We collect the popular design motifs from the existing literature and use the following set of building operators to craft the dense connectivity space:
\begin{itemize}
    \item Convolution 1$\times$1.
    \item Depthwise convolution 3$\times$3, 5$\times$5, or 7$\times$7.
\end{itemize}

Convolution 1$\times$1 and depthwise convolution are heterogeneous building operators with distinct functionality on input features: convolution 1$\times$1 learns a transformation of local input features, and depthwise convolution learns a transformation of spatial input features.
Next, we instantiate two dense connectivity spaces for ImageNet and CIFAR-10 as follows:
\begin{itemize}
    \item \noindent \textit{ImageNet Dense Connectivity Space.} Each meta-graph contains 4 stages which corresponds to the $4\times4$, $8\times8$, $16\times16$, $32\times32$ down-sampling region of an input image. In each DAG, we employ one input vertex, one output vertex, and 16 intermediate vertices assigned with one of the aforementioned convolutional building operators. We follow MobileNetV2~\cite{sandler2018mobilenetv2} for the design of stem architecture (i.e., first three blocks) and head architecture (i.e., last two blocks).
    \item \noindent \textit{CIFAR-10 Dense Connectivity Space.} Each meta-graph contains 4 stages which corresponds to the $1\times1$, $2\times2$, $4\times4$ down-sampling region of an input image. In each DAG, we employ one input vertex, one output vertex, and 16 intermediate vertices assigned with one of the aforementioned convolutional building operators. We follow ResNet~\cite{he2016deep} to design stem architecture (i.e., first block) and employ no head architecture.
\end{itemize}

Notably, we set the number of vertices to far exceed that of versatile building operators in dense connectivity design space to ensure scalability.
The dense connectivity design space is prohibitively large. Even with $N=8$ vertices, a single DAG contains $4.5\times 10^{6}$ architectures.
Consequently, an \textit{ImageNet Dense Connectivity Space} contains up to $4\times 10^{26}$ architectures, calling for an effective and efficient algorithm to demystify dense connectivity optimization thoroughly.

\section{Demystifying Dense Connectivity Search}
A dense connectivity design space contains a rich source of candidates to ensure flexibility. However, versatile building operators and the dense connectivity design space challenge the efficiency and quality of search.
CSCO incorporates two key techniques that facilitate architecture exploration in the dense connectivity design space. First, CSCO adopts Graph Isomorphism to augment architecture-performance pairs in the dense connectivity design space to boost the accuracy predictor's prediction quality without additional cost.
Second, CSCO proposes a novel search strategy, Metropolis-Hastings Evolutionary Search (MH-ES), as an in-place improvement over evolutionary search during full search space exploration via a trained predictor.
Inspired by the definition of Markov Chains, MH-ES rejects weaker samples with a lower probability and effectively evades local optimal solutions in discovery.

\begin{figure*}[t]
    \begin{center}
    \includegraphics[width=1.0\linewidth]{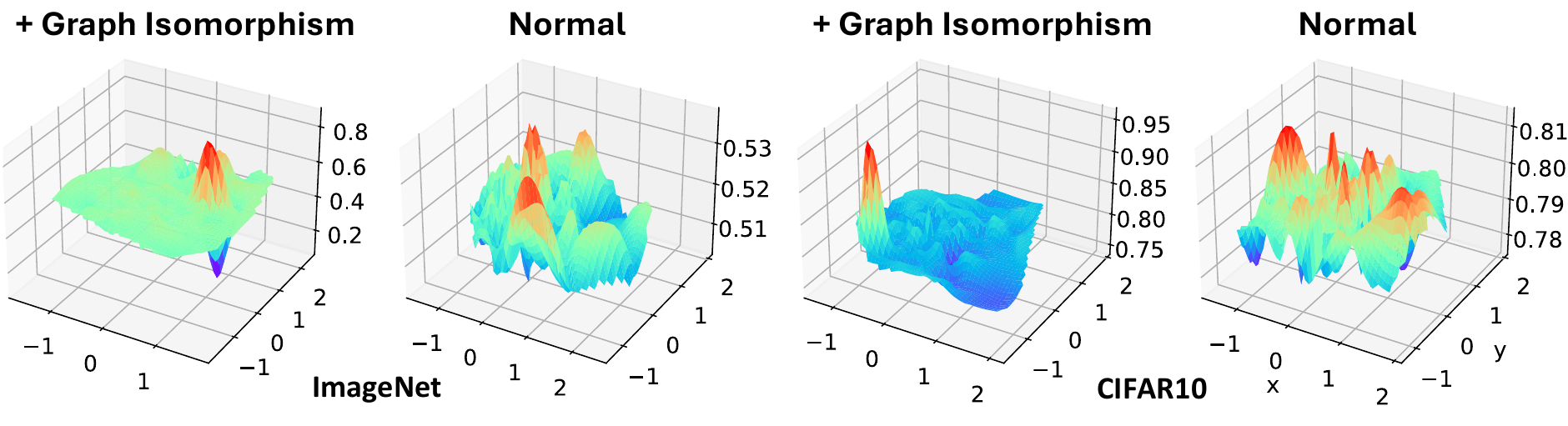}
    \caption{Accuracy surface of a performance predictor with/without Graph Isomorphism.}
    \label{fig:pd_surf}
    \end{center}
\end{figure*}

\subsection{Graph Isomorphism as Data Augmentation}
\label{sec:graph_aug}

As is discussed in Section \ref{sec:building_op}, a DAG within a dense connectivity design space contains more building operators than the number of vertices.
This provides a chance to find isomorphic structures in the dense connectivity design space and further exploit these architectures to enhance the performance of predictor-based NAS. 
We formally define the isomorphism of meta-graphs as follows:
\begin{definition}
Two meta-graphs $\mathbf{G}, \hat{\mathbf{G}}$ are isomorphic if every pair of DAG: $(\mathcal{G}^{(\cdot)}, \hat{\mathcal{G}}^{(\cdot)})$ is isomorphic. 
\end{definition}

Isomorphic meta-graphs represent the same architecture in the dense connectivity search space.
This is because isomorphic meta-graphs have an identical set of building operators and identical dense connectivity of these building operators, see Figure \ref{fig:graph_aug}. 
As a result, isomorphic meta-graphs represent the same neural architecture, leading to the same level of performance during evaluation.

Thus, we propose Graph Isomorphism to augment the architecture samples.
Graph Isomorphism conducts a valid vertex permutation to one of the DAGs within each sampled meta-graph to construct a new isomorphic meta-graph and incorporate it as a new architecture sample with no extra search cost.
These isomorphic samples can augment the architecture-performance pairs to brew a more accurate performance predictor without additional search costs.

\noindent \textbf{Prediction Surface.} We visualize the prediction surface of performance predictors with/without Graph Isomorphism in Figure \ref{fig:pd_surf}. 
Here, a higher z-axis value denotes better predictive performance on the target dataset.
Notably, Graph Isomorphism not only enhances the prediction quality of a performance predictor but also provides a smooth performance surface that eases the following search process in an ample, dense connectivity design space.

\subsection{Metropolis-Hastings Evolutionary Search}
\label{sec:mh_es}
Evolutionary Search is a popular method that efficiently explores the best architecture in predictor-based NAS. Yet, these methods may not efficiently explore our dense connectivity design space, thus suffering from the sub-optimal quality of discovered CNN architectures.
We follow the same intuition of evolutionary search and first define the mutation space as follows:
\begin{itemize}
    \item Re-sample a random edge connection for one DAG.
    \item Randomly add an edge connection for one DAG.
    \item Randomly remove an edge connection for one DAG.
\end{itemize}

Given an intermediate meta-graph with $N$ vertices and $K$ stages, the mutation space covers up to $O(N^{2K})$ possible candidate architectures, thus being prohibitively large for existing evolutionary search algorithms to explore fully. For example, (1) tremendous samples are needed to cover the good regions of the dense connectivity space and obtain the best child architecture, and (2) the complexity of prediction surface in dense connectivity design space may lead to the discovery of locally optimal solutions.
This is because evolutionary search judiciously accepts the strongest child architectures during the evolutionary process and, thus, obtains locally optimal solutions with high concentration on a specific region of the dense connectivity design space.

We are inspired by Markov Chain Monte Carlo (MCMC) optimization, especially Metropolis-Hastings Algorithm~\cite{metropolis1953equation}, extensively addressing such issues by adopting an acceptance-rejection mechanism.
Such mechanism maintains a current best solution and admits weaker solutions with an acceptance-rejection probability $AC$, defined as follows:
\begin{equation}
    AC = \min(1, \exp{((score'-score)/T)}),
\end{equation}
Where $score$ denotes the score (e.g., predicted performance of architectures) for a weaker solution, $score'$ denotes the score of the current best solution, and $T$ denotes the temperature. 
The acceptance-rejection probability is proportional to the gap between the weaker solution and the current best solution. 
As a result, the optimization process may not greedily stick to the current best solution and thus have better potential to avoid locally optimal solutions.

\begin{figure*}[t]
    \begin{center}
    \includegraphics[width=1.0\linewidth]{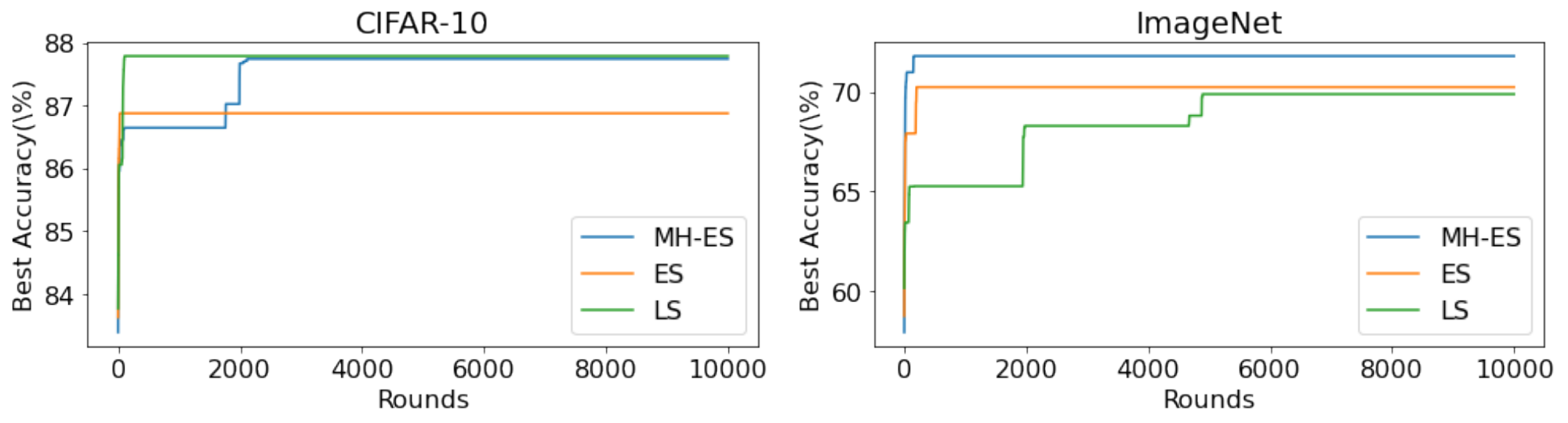}
    \vspace{-2em}
    \caption{The search progress and predicted accuracy of discovered architectures via MH-ES compared to ES and RS baselines.}
    \label{fig:mhes_curve}
    \end{center}
\end{figure*}

Thus, we propose Metropolis-Hastings Evolutionary Search (MH-ES) as an alternative to existing evolutionary search algorithms on dense connectivity design space.
MH-ES allows the discovery of better architectures within the large dense connectivity design space.
MH-ES maintains the best parent architecture, which is initialized among $P_0$ randomly sampled candidate architectures in the initial population. 
Then, child architectures are obtained by randomly mutating one of the stages (i.e., DAGs) in the parent architecture (i.e., meta-graph).
Each evolution round selects the best child architecture in the current population as a weaker solution. The aforementioned MH acceptance-rejection ratio $AC$ is applied to update the current best architecture.
The proposed MH-ES generalizes to local search when $T \to 0$ and evolutionary search when $T \to \infty$. 
MH-ES also adopts a cosine simulated annealing~\cite{pincus1970letter} of the temperature to eliminate locally optimal solutions at early evolutionary rounds. 

\noindent \textbf{Optimization Curve of MH-ES.}
We compare the optimization curve of MH-ES versus Evolutionary Search (ES)~\cite{goldberg1991comparative} and Local Search (LS)~\cite{friggstad2019local} to demonstrate its effectiveness and efficiency.
Figure \ref{fig:mhes_curve} depicts the optimization curve of accuracy on top-performing CNN architectures for both CIFAR-10 and ImageNet.
On a small-scale CIFAR-10 dense connectivity design space, MH-ES performs on-par as local search yet significantly outperforms evolutionary search by $\sim 0.01$. This is greatly attributed to the capability of MH-ES to evade locally optimal solutions during architecture exploration over the dense connectivity design space.
On large-scale ImageNet dense connectivity design space, MH-ES significantly edges other search algorithms, highlighting its efficiency and effectiveness in exploring dense connectivity design space.


\section{Experiments} 
Following all NAS methods focusing on connectivity, we apply CSCO to obtain promising CNN architectures within mobile computation regimes on CIFAR-10/ImageNet-1K~\cite{deng2009imagenet} classification over \textit{CIFAR-10 Dense Connectivity Space} / \textit{ImageNet Dense Connectivity Space}.

\begin{table}[b]
    \begin{center}
    \caption{CIFAR-10 evaluation of best models.}
    \vspace{-0.5em}
    \scalebox{0.8}{
    \begin{tabular}{|c|c|c|c|}
    \hline
     \multirow{2}{*}{\textbf{Architecture}}   & \textbf{Test} &  \multirow{2}{*}{\textbf{Params (M)}} & \textbf{Search Cost} \\
     & \textbf{Error} (\%) & & \textbf{(GPU Days)} \\
    \hline\hline
        WRN-28-10~\cite{zagoruyko2016wide} & 4.17 & 36.5 & - \\
        DenseNet-BC~\cite{huang2017densely} & 3.46  & 25.6 &  - \\
        \hline \hline
        PNAS~\cite{liu2018progressive} & 3.41{\tiny$\pm$0.09} & 3.2 & - \\
        AmoebaNet-A~\cite{real2019regularized} & 3.34{\tiny$\pm$0.06} & 3.2 & 3150 \\
        DARTS (1st-order)~\cite{liu2018darts} & 3.0{\tiny$\pm$0.14} & 3.3 &  4 \\
        GDAS~\cite{dong2019searching} & 2.93 & 3.4 & 0.3 \\
        CSCO (Ours) &  \textbf{2.82} & \textbf{3.1} & 4 \\
        \hline
    \end{tabular}
    }
    \label{tab:cifar_results}
    \end{center}
\end{table}

\subsection{CSCO Setup}
We first elaborate on the search settings on CSCO, including search space configuration and MH-ES guided by a trained neural predictor.
Then, we discuss the evaluation settings of CSCO over a dense connectivity design space. 

\noindent \textbf{Search Space Settings.}
We set a search budget of 4 GPU days and employ a fixed assignment of the building operators in all DAGs of the meta-graph. This enhances the reproducibility of our methods.
We employ a large-scale dense connectivity design space with $N=18$ vertices, where vertex $1, 5, 9, 13$ are assigned with convolution 1$\times$1, vertices $2, 6, 10, 14$ are assigned with depthwise convolution 3$\times$3, vertex $3, 7, 11, 15$ are assigned with depthwise convolution 5$\times$5, and vertex $4,8,12,16$ are assigned with depthwise convolution 7$\times$7.
We define vertex 0/17 as the input/output vertex.

\noindent \textbf{Predictor Training and MH-ES.} We train the above MLP performance predictor on the sampled architecture-performance pairs for 300 epochs with batch size 128, initial learning rate 0.1, and an L2 weight decay of 1e-4 for ImageNet. During MH-ES, we employ an initial population of 4096 to ensure the discovery of a good parent architecture. We proceed with 10K rounds of MCMC optimization with 96 child architectures sampled and evaluated in each round. We set the sensitivity parameter to 0.001 for the best solution in the dense connectivity design space.

\begin{table*}[t]
    \caption{ImageNet-1K results. All models use 224$\times$224 input under mobile settings.}
    \vspace{-1em}
    \begin{center}
    \scalebox{1.0}{
    \begin{tabular}{|c|cc|c|c|c|c|}
    \hline
     \multirow{2}{*}{\textbf{Architecture}} & \multicolumn{2}{|c|}{\textbf{Test Err.(\%)}}  &  \textbf{MACs} & \textbf{Params}   & \textbf{Search Cost} \\
      &  \textbf{top-1} & \textbf{top-5} &  \textbf{(M)} & \textbf{(M)}  & \textbf{(GPU days)} \\
    \hline \hline
        CondenseNet~\cite{huang2018condensenet} & 26.2 & 8.3 & 529 & 4.8 & N/A \\
        MobileNetV2 1.4~\cite{sandler2018mobilenetv2}& 25.3 & 7.5  & 585 & 6.9 & N/A \\
        \hline
        ProxylessNAS-G~\cite{cai2018proxylessnas}& 25.4 &  7.8  & 320 & 4.1 & 8.33 \\
        MnasNet-A1~\cite{tan2019mnasnet}& 24.8 & 7.5  & 312 & 3.9 & 1.7K \\
        DARTS~\cite{liu2018darts}&  26.7 & 8.7 & 574 & 4.7 & 4 \\
        NASNet-A~\cite{zoph2018learning} & 26.0 & 8.4 & 564 & 5.3 & 2K \\
        RandWire-WS~\cite{Xie_2019_ICCV} & 25.3{\tiny$\pm$0.25}& 7.8{\tiny$\pm$0.15}  & 583{\tiny$\pm$6.2} & 5.6{\tiny$\pm$0.1}  & N/A \\
        MiLeNAS~\cite{he2020milenas}& 24.7 & 7.6 & 584 & 5.3 & 0.3 \\
        PC-DARTS~\cite{xu2019pc} & 24.2 & 7.3 & 597 & 5.3 & 3.8 \\
        TopoNAS~\cite{huang2020explicitly} & 24.1 & 7.2 & 598 & 5.3 & 6.2 \\
        GAEA + PC-DARTS~\cite{li2020geometry} & 24.0 & 7.3 & N/A & 5.6 & 3.8 \\
        DenseNAS~\cite{fang2020densely} & 23.9 & - & 479 & - & 2.67 \\
        CSCO (Ours) & \textbf{23.3}  & \textbf{6.7} & 598 & 5.7 & 8 \\
        \hline
    \end{tabular}}
    \end{center}
    \label{tab:imagenet_results}
\end{table*}

\noindent \textbf{Evaluation Settings.} 
The outcome of CSCO leads to a pool of candidate CNN architectures for both CIFAR-10 and ImageNet, respectively. 
We simply evaluate the top-5 models on CIFAR-10/ImageNet proxy dataset for 20/10 epochs and scale the best model to 600M Multiply-Accumulates (MACs) mobile computation budget. 

\subsection{CIFAR-10 Experiments}
We first evaluate each component of the CSCO paradigm and then proceed to evaluate the top-performing architecture discovered on CIFAR-10.

\noindent \textbf{Evaluating Best CIFAR-10 Model.}
In CIFAR-10, we follow DARTS-series architectures~\cite{liu2018darts,xu2019pc}) and stack 6 building cells in each stage to construct the final CNN architecture. To match the number of parameters reported in the DARTS-series paper, we apply a width multiplier to scale up the candidate networks to ensure a fair comparison with existing state-of-the-art.

We follow the DARTS protocol to train the best network. 
Specifically,
we train the best network discovered by CSCO on 50K CIFAR-10 training data from scratch for 600 epochs with batch size 96. 
We employ an initial learning rate of 0.025 with a cosine learning rate schedule~\cite{loshchilov2017sgdr}. 
Following DARTS series works, we employ Dropout~\cite{hinton2012deep}, DropPath~\cite{geman1984stochastic} and Cutout~\cite{devries2017improved} with a L-2 weight-decay of 3e-4 to combat overfitting. 

The key results of CSCO on the CIFAR-10 dataset are summarized in Table \ref{tab:cifar_results}. 
CSCO outperforms SMBO-based PNAS and EA-based AmoebaNet by up to 0.42\%. Compared with DARTS and GDAS, CSCO achieves up to 0.2\% better accuracy within a reasonable 4 GPU day search cost.

\subsection{ImageNet Classification}
We evaluate the best architectures crafted by CSCO on ImageNet by training them from scratch using the same training protocol as previous works.
We compare the accuracy versus various metrics such as Multiply-Accumulates (MACs) with hand-crafted and NAS-crafted models. 
We train the best-discovered model on 1.28M ImageNet-1K training data from scratch for 450 epochs with batch size 768. We employ an initial learning rate of 0.6 with cosine learning rate schedule~\cite{loshchilov2017sgdr}. 
Following DARTS series works, we employ Inception pre-processing~\cite{Szegedy_inceptionv4}, Dropout~\cite{hinton2012deep}, Drop Path~\cite{geman1984stochastic}, an L2 weight-decay of 1e-5.

Table \ref{tab:imagenet_results} demonstrates the critical results of CSCO on the ImageNet-1K validation set within the mobile computation regime (i.e., $\le$ 600M MACs).
CSCO outperforms Condensenet~\cite{huang2018condensenet} by 3\% higher accuracy, demonstrating its superiority over prior art with dense connectivity of building operators in CNN architectures.
CSCO outperforms SOTA hand-crafted models MobileNetV2 1.4 by 2.0\% higher top-1 accuracy with similar MACs, where the latter one is crafted via manual architecture engineering.
Compared to existing NAS works that emphasize dense connectivity~\cite{xu2019pc,fang2020densely,he2020milenas,huang2020explicitly,li2020geometry}, CSCO achieves $\sim0.6\%$ accuracy gain under the mobile computation regime with comparable parameter consumption.
Despite having a sizeable dense connectivity design space, CSCO maintains a reasonable search cost of 8 GPU days thanks to Graph Isomorphism, which boosts sample efficiency.
Finally, CSCO also achieves competitive performance compared to existing block-based NAS methods~\cite{tan2019mnasnet,cai2018proxylessnas} under a well-designed MobileNetV2-like search space, demonstrating the potential of optimizing dense connectivity to seek high-performing CNN architectures in the future.

\section{Discussion}

CSCO employs two key components in dense connectivity search space: Graph Isomorphism and Metropolis-Hastings Evolutionary Search (MH-ES). 
In this section, we discuss the individual contribution of Graph Isomorphism and MH-ES towards better dense connectivity in discovery.

\begin{table*}[t]
    \begin{center}
    \caption{Evaluation of Graph Isomorphism and MH-ES on CIFAR-10.}
    \vspace{-0.5em}
    \begin{tabular}{|c|c|c|c|c|}
         \hline
         \textbf{Graph Isomorphism?} & \textbf{Search Strategy} & \textbf{Best Acc.} & \textbf{Mean Acc.} & \textbf{Std. Acc.} \\
         \hline
         - & Random Search (RS) & 91.52 & - & - \\
         & MH-ES & 92.26 & 92.11 & 0.10 \\
         \checkmark & MH-ES & 92.55 & 92.39 & 0.11 \\
         \checkmark & Local Search (LS) & 92.33 & 92.25 & 0.05\\
         \checkmark & Evolutionary Search (ES) & 92.49 & 92.07 & 0.23 \\
         \hline
    \end{tabular}
    \label{tab:ablation_graphaug_mhes}
    \end{center}
\end{table*}

\subsection{Ablation Studies}

Due to the time-consuming process of complete architecture evaluation from scratch, we adopt a simple training protocol to evaluate the top models discovered by different Graph Isomorphism and search strategies. Table \ref{tab:ablation_graphaug_mhes} demonstrates the detailed evaluation result, including the accuracy of best-performing architectures and the statistics on top-5 models to reflect the stability of the proposed method. 
Here, we compute all accuracies using top models selected by the trained predictor.
Notably, each model contains only 3 building cells in each stage, and the building operator adopts 16, 32, and 64 filters, respectively, within each stage.
We can see that the combination of MH-ES and Graph Isomorphism yields up to a 0.3\% accuracy gain on the mean accuracy of top-performing models while achieving the best top-performing architecture among other baseline methods.

\subsection{Ranking with Graph Isomorphism}
We analyze the performance predictor trained with/without Graph Isomorphism.
Before Graph Isomorphism, we sample $\sim$ 800 samples from the dense connectivity design space for both CIFAR-10 and ImageNet benchmark and augment $10\times \sim 12 \times$ more samples via Graph Isomorphism without extra computation cost.
We split all architecture-performance pairs into 85\% training pairs and 15\% testing pairs. 
We utilize a Multi-Layer Perceptron (MLP) performance predictor to map architectures (i.e., the union of adjacency matrices in each DAG within a meta-graph) to their predicted performance (i.e., evaluated accuracy).
We delicately train the performance predictor on the training split with/without Graph Isomorphism.

\noindent \textbf{Prediction Quality.} 
We measure the prediction quality of neural predictors on the testing architecture-performance pairs via two famous ranking metrics: Pearson's $\rho$ and Kendall's $\tau$ in Table \ref{tab:graphaug}. Here, all ranking coefficients are computed on the testing pairs, which are not utilized to train the performance predictor.

\begin{table}[t]
\caption{Evaluation of prediction quality with/without Graph Isomorphism (GI).}
\vspace{-1em}
\begin{center}
    \scalebox{0.74}{
    \begin{tabular}{|c|c|c|c|c|}
    \hline
    \textbf{Dataset} & \textbf{GI?} & \textbf{Pearson's $\rho$} & \textbf{Kendall's $\tau$} & \textbf{Mean-Squared Error} \\
    \hline
    \multirow{2}{*}{\textbf{ImageNet}} &  & 0.342 & 0.215 & $5\times10^{-4}$\\
    & \checkmark & 0.972 & 0.904 & $4\times 10^{-5}$ \\
    \hline
    \multirow{2}{*}{\textbf{CIFAR-10}} & & 0.581 & 0.428 & $4 \times10^{-4}$ \\
    & \checkmark & 0.946 & 0.874 & $6 \times 10^{-5}$\\
    \hline
    \end{tabular}
    }
\end{center}
\label{tab:graphaug}    
\end{table}

Using Graph Isomorphism, the prediction ranking quality (i.e., Kendall's $\tau$) significantly increases from 0.215 to 0.904 on ImageNet and from 0.428 to 0.874 on CIFAR-10. In addition, the ranking evaluation reveals that more DAGs in the meta-graph lead to poorer sample efficiency in predictor-based NAS and thus lead to a more challenging search process on top-performing models. In contrast, Graph Isomorphism judiciously incorporates isomorphic graph transformation into each independent DAG in the meta-graph, yielding an even more significant prediction quality improvement on large-scale \textit{ImageNet Dense Connectivity Space} over small-scale \textit{CIFAR-10 Dense Connectivity Space}.

\section{Conclusion} 
We propose CSCO, a novel paradigm that allows flexible exploration of the dense connectivity of building operators and innovates building cells in CNN architectures.
CSCO aims to seek the optimal building cells of CNN architectures represented by Directed Acyclic Graphs (DAGs), containing rich sources of dense connectivity of versatile building operators to cover CNN architecture designs flexibly.
CSCO crafts a dense connectivity space to fabricate the building cells of the CNN architectures and further leverages a performance predictor to obtain the best dense connectivity.
To enhance the reliability and quality of prediction, we propose Graph Isomorphism as data augmentation to boost sample efficiency and Metropolis-Hastings Evolutionary Search (MH-ES) to efficiently explore dense connectivity space and evade locally optimal solutions in CSCO.
Experimental on ImageNet demonstrates $\sim$ 0.6\% accuracy gain over other NAS-crafted dense connectivity designs under mobile computation regime.

\noindent \textbf{Acknowledgement.} This project is partly supported by NSF 2112562, ARO W911NF-23-2-0224, NSF CAREER-2305491, and NSF 2148253.

{
    \small
    \bibliographystyle{ieeenat_fullname}
    \bibliography{main}
}
\end{document}